\documentclass[preprint,12pt]{elsarticle}

\usepackage{amssymb}
\usepackage{amsmath}
\usepackage{booktabs}
\usepackage{tabularx}
\usepackage{graphicx}
\usepackage{float}
\usepackage{etoolbox}
\usepackage{hyperref}
\usepackage{xurl}
\usepackage{pgfplots}
\pgfplotsset{compat=1.18} 
\usepgfplotslibrary{groupplots}
\usepackage{algorithm}
\usepackage{algpseudocode}
\usepackage{tikz}
\usetikzlibrary{positioning,arrows.meta,shapes.multipart,calc,patterns,matrix,backgrounds,fit,decorations.pathreplacing}

\journal{Neurocomputing}
\makeatletter
\patchcmd{\pprintMaketitle}{\elsaddress\par\vskip36pt}{\elsaddress\par\vskip18pt}{}{\typeout{main.tex: gap A patch failed}}
\patchcmd{\pprintMaketitle}{\Large\@title\par\vskip18pt}{\Large\@title\par\vskip14pt}{}{\typeout{main.tex: gap B patch failed}}
\patchcmd{\pprintMaketitle}{\hrule\vskip12pt}{\hrule\vskip10pt}{}{\typeout{main.tex: gap D patch failed}}
\patchcmd{\pprintMaketitle}{\unvbox\absbox\par\vskip10pt}{\unvbox\absbox\par\vskip8pt}{}{\typeout{main.tex: gap E patch failed}}
\patchcmd{\pprintMaketitle}{\unvbox\keybox\par\vskip10pt}{\unvbox\keybox\par\vskip8pt}{}{\typeout{main.tex: gap F patch failed}}
\makeatother

\begin{document}
\begin{frontmatter}

\title{Beyond Causal Masks: Auditing and Localizing Hidden Future-Token Leakage in Adaptive Sequence Models}

\author[inst1]{Huiting Liu}
\author[inst1,inst2]{Jinsong Wu\corref{cor1}}
\affiliation[inst1]{
  organization={School of Artificial Intelligence, Guilin University of Electronic Technology},
  city={Guilin},
  postcode={541004},
  country={China}
}
\affiliation[inst2]{
  organization={Department of Electrical Engineering, University of Chile},
  city={Santiago},
  postcode={8370451},
  country={Chile}
}
\cortext[cor1]{Corresponding author. Email address: \texttt{wujs@ieee.org}}

\begin{abstract}
Hidden future-token leakage can create false architectural gains: an autoregressive model may appear substantially better under teacher-forced evaluation while its free-running generation deteriorates. We investigate this failure in an adaptive residual-routing decoder whose sequence-level routing summary exposes prefix positions to future tokens despite a correct causal attention mask. An anomalous 24-layer checkpoint reduces validation loss from the baseline's 3.5958 to 1.4259, yet it fails a prefix-invariance audit and has substantially higher reference-model perplexity during generation. Layer-wise diagnostics and implementation inspection trace the violation to a full-sequence summary broadcast through the dynamic routing-query path. Two independently trained 12-layer causal repairs eliminate detectable leakage and return validation loss and reference perplexity to the control scale, removing the large apparent advantage. As diagnostic tools, we use the established prefix-invariance contract, maximum prefix-output discrepancy (LeakScore), complementary behavioral metrics, and layer-wise profiles. Supporting experiments cover six synthetic implementation pairs spanning five module families, clean pretrained controls, controlled fault injection, and cache consistency across numerical precisions. These results illustrate how hidden future-token access can produce misleading claims of architectural progress and why lower teacher-forced loss should be interpreted together with whole-model causal checks and free-running evaluation.
\end{abstract}


\end{frontmatter}

\section{Introduction}
In decoder-only Transformers, causal self-attention restricts the output at position $t$ to information from the prefix $x_{1:t}$. The same restriction must hold for every operation contributing to that output. Sequence-level pooling, temporal convolution, state aggregation, routing, and cross-time normalization can create additional information paths outside of masked attention. Prior work on linear attention and recurrent-style sequence compression likewise makes autoregressive behavior depend on prefix-consistent state aggregation \cite{katharopoulos2020transformers}. Adaptive computation and selective state updates further motivate checking the complete dependency structure \cite{raposo2024mixture,gu2024mamba}.

We use \emph{hidden future-token leakage} to describe suffix influence on prefix outputs through an auxiliary operation outside standard masked self-attention, despite a correct attention mask. Such access can lower teacher-forced validation loss without improving valid autoregressive prediction. Broader studies of leakage have shown how invalid information access can inflate empirical performance claims \cite{kapoor2023leakage}; here we examine this risk within the model's computation. The practical concern is a \emph{false architectural gain}: an apparent improvement that depends on information unavailable during causal generation.

\begin{figure}[h]
\centering
\includegraphics[width=0.65\textwidth]{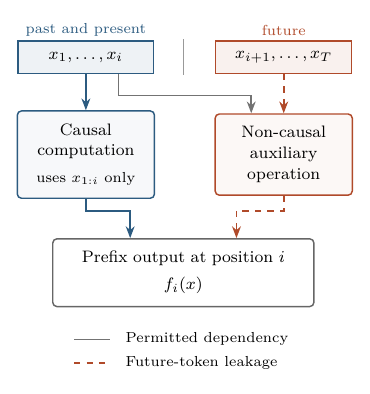}
\caption{Mechanism of hidden future-token leakage. Solid paths represent permitted dependencies on past and present inputs, while the orange dashed path highlights a forbidden dependency on future inputs through a non-causal auxiliary operation. This is a schematic dependency diagram rather than a particular architecture: pooling, temporal mixing, state aggregation, routing, and cross-time normalization can introduce such paths when implemented non-causally.}
\label{fig:mechanism}
\end{figure}

Figure~\ref{fig:mechanism} illustrates the dependency structure. At position $i$, the causal path uses only $x_{1:i}$, while a non-causal auxiliary operation introduces an additional dependency on $x_{i+1:T}$. The auxiliary operation may also use past and present inputs; the violation is the future-to-prefix dependency, not the presence of the auxiliary module itself.

Our motivation comes from a follow-up implementation of CDR-AttnRes (Content-Conditioned Depth Routing for Attention Residuals), a residual-routing decoder variant inspired by Attention Residuals \cite{kimiteam2026attentionresiduals}. A full-sequence summary was broadcast as a low-rank correction to the depth-routing query. An unusually low teacher-forced validation loss suggested a strong improvement, but causal auditing detected leakage and free-running generation had substantially higher reference-model perplexity than the controls. Localization and code inspection identified the query path; independently trained 12-layer repairs removed detectable leakage and the large apparent advantage. Section~\ref{sec:cdr_case} distinguishes the anomalous 24-layer trigger from the 12-layer repair comparisons.

Our diagnostic framework uses the established \emph{prefix-invariance} principle: changing the suffix should leave shared-prefix outputs unchanged. We denote the maximum prefix-output discrepancy by \emph{LeakScore}, supplement it with behavioral metrics, and inspect layer-wise profiles to guide implementation review. Controlled module comparisons and pretrained-model checks support interpretation of these diagnostics.

The main contributions are threefold. First, we document \emph{a false architectural gain in a real adaptive architecture}: a large teacher-forced loss advantage accompanies detectable leakage and generation degradation, and disappears once the non-causal path is repaired. Second, we provide \emph{cross-module diagnostic evidence} across six implementation pairs spanning five module families, together with clean controls and fault-injected pretrained models. Third, we use \emph{layer-wise localization to support root-cause analysis}: FirstLeakLayer identifies the first monitored layer with detectable suffix influence and guides inspection of the CDR query path. Prefix invariance and LeakScore serve as diagnostic tools rather than a new causal contract.

Sections~\ref{sec:problem_definition}--\ref{sec:method} define the problem and audit. Sections~\ref{sec:synthetic_experiments}--\ref{sec:real_model_audits} present supporting experiments, followed by the CDR case in Section~\ref{sec:cdr_case} and practical implications in Section~\ref{sec:discussion}.

\section{Related Work}
\label{sec:related_work}

\paragraph{Leakage and empirical reliability}
Data leakage can produce overoptimistic empirical conclusions \cite{kapoor2023leakage}, while overlap between benchmark content and training data complicates evaluation \cite{xu2024benchmark,dodge2021c4}. These concerns arise at the dataset or evaluation-protocol level. Our focus is an internal dependency violation: future tokens influence prefix computation even when the dataset split and attention mask are correct.

\paragraph{Causality contracts}
FBS describes a no-leakage unit test comparing shared-prefix logits under different suffixes by $\ell_\infty$ discrepancy, together with preview-cache consistency checks \cite{fbs2026}. TorchLean formalizes causality-related contracts for attention and state-space sequence models, including prefix preservation, cache-position alignment, and scan consistency. Its diagnostic examples also address normalization state and padding conventions \cite{torchlean2026}. We build on the established prefix-invariance principle. Our empirical emphasis is a documented case in which a false gain disappears under causal repair, supported by cross-module diagnosis and layer-wise localization.

\paragraph{Sequence mixing beyond standard attention}
Masked self-attention enforces causal token access in decoder-only Transformers \cite{vaswani2017attention}; segment-level and sparse-window variants support longer contexts \cite{dai2019transformerxl,beltagy2020longformer}. Linear attention and recurrent-form Transformers aggregate history into prefix states \cite{katharopoulos2020transformers}. Causal convolutions prevent access to future inputs \cite{vandenoord2016wavenet}, and state-space models use structured or selective state updates \cite{gu2022s4,gu2024mamba,dao2024mamba2}. RetNet, RWKV, and Hyena explore retention, recurrent formulations of linear attention, and long-convolution/gating mechanisms, respectively \cite{sun2023retnet,peng2023rwkv,poli2023hyena}. Their different computational forms motivate examining the causal structure of state aggregation and temporal mixing beyond the attention mask. Our controlled violations concern the auxiliary implementations specified in the experiments, rather than a claim that these cited architectures are non-causal.

\paragraph{Adaptive computation}
Mixture-of-experts models route tokens to selected experts \cite{shazeer2017moe,fedus2022switch}. Expert-choice routing assigns tokens to experts by ranking across sequence positions, so a position's assignment can depend on later positions \cite{zhou2022expertchoice}. Expert threshold routing addresses causal routing and computation allocation \cite{sun2026etr}, while Mixture-of-Depths allocates computation across tokens and layers \cite{raposo2024mixture}. These works make the information available to routing decisions a relevant implementation concern.

\paragraph{Testing, verification, and residual routing}
Differential and whitebox testing detect behavioral inconsistencies \cite{pei2017deepxplore}, whereas certified robustness methods derive provable bounds under specified perturbations \cite{xu2020lirpa}, and formalization frameworks express semantic contracts \cite{torchlean2026}. Our behavioral audit requires access to position-wise outputs, and layer-wise localization additionally requires access to the monitored hidden states; compatibility with arbitrary checkpoints is not assumed. Attention Residuals provides the residual-aggregation setting for our follow-up CDR implementation \cite{kimiteam2026attentionresiduals}. The case concerns the full-sequence query correction added in that implementation.

\section{Problem Definition}
\label{sec:problem_definition}

We consider a decoder-only autoregressive sequence model $f$ operating on an input sequence
\[
x = (x_1,x_2,\ldots,x_T).
\]
Causal correctness requires that, for any two inputs $x$ and $x'$ with $x_{1:t}=x'_{1:t}$, the outputs satisfy $f(x)_{1:t}=f(x')_{1:t}$. This requirement concerns the complete computation of $f$, including auxiliary operations as well as attention.

We use \emph{hidden future-token leakage} to denote a violation of this condition mediated by an auxiliary operation outside masked self-attention, while the attention mask itself is causal. The term ``hidden'' refers to a forbidden dependency that is not visible in the attention mask.

This notion should be clearly distinguished from ordinary train/test leakage or benchmark contamination in machine learning \cite{kapoor2023leakage,xu2024benchmark}. In conventional train/test leakage, the main problem is that evaluation data, labels, or benchmark content improperly overlaps with the training process, which leads to overoptimistic performance estimates at the dataset or protocol level. By contrast, hidden future-token leakage is an \emph{intra-model causal violation}. The training and validation split may be entirely correct, yet the model can still behave non-causally at inference time because suffix information flows into prefix computation through internal modules. In other words, ordinary leakage concerns whether the model has seen the wrong data, whereas hidden future-token leakage concerns whether the model computes the wrong dependency structure.

Our audit examines which auxiliary operations violate this contract, how strongly their outputs depend on suffix perturbations, and where that dependence first becomes detectable. We use the maximum prefix-output discrepancy, denoted LeakScore, together with complementary behavioral metrics and layer-wise profiles.

\section{Audit Methodology}
\label{sec:method}

To detect hidden future-token leakage in a module-agnostic way, we use a simple causal auditing principle based on \emph{prefix invariance}. The core idea is that, for a causally correct autoregressive model, changing only the suffix of an input sequence should not affect the model outputs on the shared prefix.

\subsection{Prefix-Invariance Criterion}

Let $x$ and $x'$ be two input sequences such that
\[
x_{1:t} = x'_{1:t}, \qquad x_{t+1:T} \neq x'_{t+1:T}.
\]
If the model $f$ is causally correct, then its outputs on the shared prefix should remain unchanged up to numerical tolerance:
\[
f(x)_{1:t} \approx f(x')_{1:t}.
\]
This is the criterion used throughout the paper. The prefix is held fixed and the suffix is perturbed. A discrepancy exceeding a suitably calibrated numerical tolerance, with other sources of variation controlled, is evidence of suffix-to-prefix influence. Because this test is defined only in terms of input-output behavior, it can be applied uniformly across a wide range of sequence modules, regardless of their internal implementation.

\subsection{LeakScore}

We summarize the discrepancy between the two prefix outputs induced by suffix perturbation using the following maximum-difference statistic, which we denote \emph{LeakScore}:
\[
\mathrm{LeakScore}(f;x,x',t)
= \max_{i \le t}\left\|f_i(x)-f_i(x')\right\|_{\infty}.
\]
This quantity measures the largest absolute output deviation over all prefix positions and output coordinates. For language models, the output coordinates are vocabulary logits. A score exceeding the numerical tolerance indicates detectable suffix-to-prefix influence for the tested pair; an undetected dependency may still exist for other inputs or below the tolerance.

For completeness, one may also consider an averaged version,
\[
\mathrm{LeakScore}_{\mathrm{mean}}(f;x,x',t)
= \frac{1}{td}\sum_{i=1}^{t}\left\|f_i(x)-f_i(x')\right\|_{1},
\]
where $d$ is the output dimension (vocabulary size for logits). This is the elementwise mean absolute difference reported as Mean diff; batches are additionally averaged over examples. LeakScore determines the pair-level verdict, while the mean discrepancy summarizes the overall perturbation magnitude. We also report the top-1 flip rate (the fraction of prefix positions whose highest-scoring token changes) and the affected-token rate (the fraction whose maximum coordinate difference exceeds $\epsilon$). The normalized maximum divides LeakScore by the maximum absolute reference-prefix logit plus $10^{-12}$; mean Jensen--Shannon (JS) divergence compares the two softmax distributions and is averaged over prefix positions. These auxiliary metrics characterize the discrepancy but do not replace the absolute-tolerance verdict.

\subsection{Audit Procedure}

Algorithm~\ref{alg:prefix_invariance} summarizes the audit procedure. We construct a pair of sequences with the same prefix and different suffixes, evaluate both with the same model parameters and runtime settings, and compare their prefix outputs. Dropout is disabled and other stochastic effects are controlled so that input perturbation is the intended source of variation.

\begin{algorithm}[t]
\caption{Prefix-invariance audit procedure used throughout this work.}
\label{alg:prefix_invariance}
\begin{algorithmic}[1]
\Require model $f$, sequence pair $(x,x')$, prefix index $t$, tolerance $\epsilon$, with $x_{1:t}=x'_{1:t}$ and $x_{t+1:T}\neq x'_{t+1:T}$
\Ensure audit outcome: \textbf{pass} or \textbf{fail}
\State Compute $y=f(x)$ and $y'=f(x')$
\State Extract prefix outputs $y_{1:t}$ and $y'_{1:t}$
\State Compute $\mathrm{LeakScore}(f;x,x',t)=\max_{i\le t}\|y_i-y'_i\|_{\infty}$
\If{$\mathrm{LeakScore}(f;x,x',t)>\epsilon$}
    \State report \textbf{fail}
\Else
    \State report \textbf{pass}
\EndIf
\end{algorithmic}
\end{algorithm}

\begin{figure}[H]
\centering
\includegraphics[width=0.65\textwidth]{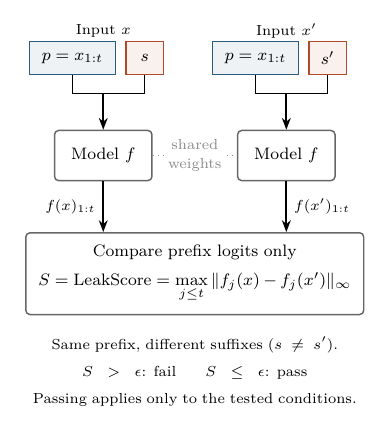}
\caption{Prefix-invariance audit. Two inputs share a prefix $p=x_{1:t}$ but have different suffixes $s$ and $s'$. Both are evaluated by the same model under the same evaluation settings, with stochastic effects controlled. LeakScore is the largest absolute prefix-logit difference across positions and vocabulary entries. A score above the numerical tolerance $\epsilon$ indicates an audit failure; passing provides evidence only for the tested inputs and conditions.}
\label{fig:prefix_invariance}
\end{figure}

Figure~\ref{fig:prefix_invariance} illustrates the test: $x=(p,s)$ and $x'=(p,s')$ have the same prefix $p=x_{1:t}$, and only the shared-prefix logits are compared. The score $S$ in the figure is $\mathrm{LeakScore}(f;x,x',t)$ defined above. A passing verdict applies only to the tested inputs and execution conditions.

In practice, the two sequences can be generated by sampling a base sequence and replacing only the suffix segment after position $t$. The same procedure applies to logits, hidden states, or other position-wise model outputs, as long as the compared quantity is the object whose autoregressive causality we wish to audit.

\subsection{Tolerance and Numerical Robustness}

Because the test compares floating-point outputs, a small numerical tolerance is necessary even for causally correct implementations. In our experiments, we use
\[
\epsilon = 10^{-5}
\]
as the default threshold. This choice is appropriate for our toy auditing setup in standard float32 evaluation and provides a stable separation between causal and non-causal modules. As shown later in the robustness checks, our main conclusions are unchanged when the tolerance is varied over $\{10^{-6},10^{-5},10^{-4}\}$: causally repaired variants remain near numerical zero, whereas non-causal variants consistently yield LeakScore values far above this range. Thus, the observed separation is not an artifact of a particular threshold choice.

\subsection{Generation Diagnostics}

In the CDR checkpoint case study, audit verdicts are compared with open-ended generation diagnostics. The reference-model perplexity $\mathrm{ppl}_{\mathrm{ref}}$ is computed using a fixed GPT-2 medium reference LM. The frozen evaluation code scores the saved text, including the prompt and generated continuation, in windows of at most $1024$ tokens with stride $512$. For each sample, it exponentiates the token-count-weighted mean window negative log-likelihood; the reported value is the arithmetic mean of these sample perplexities over the 50 prompts. Higher perplexity means the reference model assigns lower likelihood to the text. This is a model-dependent diagnostic of generation quality, not a direct measure of semantic coherence or a human evaluation. The text is re-encoded with the reference tokenizer. Sampling entropy $H=-\sum_v p_v\log p_v$ is measured in nats from the distribution after temperature scaling and top-$k$ truncation, averaged over generated steps and then prompts. Distinct-2 is the proportion of unique bigrams, and Repeat-3 is the proportion of trigram occurrences belonging to repeated trigrams; both are computed on whitespace-separated units of the saved prompt-inclusive text. Bootstrap intervals quantify uncertainty in the prompt-averaged perplexity.

\section{Synthetic Auditing Experiments}
\label{sec:synthetic_experiments}

We next use the prefix-invariance test to audit a set of synthetic sequence modules beyond masked attention. The goal of these experiments is not to maximize task performance, but to isolate whether representative auxiliary components can introduce hidden future-token leakage under controlled conditions. We construct causally repaired and non-causal counterparts for each module family and evaluate them under the same auditing procedure described in Section~\ref{sec:method}.

\subsection{Experimental Design}

Our synthetic audit covers six implementation pairs across five module families: pooling (two pairs), temporal convolution, linear attention, dynamic routing, and cross-time normalization. Each pair contains one non-causal implementation and one causally repaired counterpart, giving twelve implementations in total. The selected families are intended to span several common ways in which future information may re-enter prefix computation, including sequence-level summary broadcast, temporal mixing, full-sequence state aggregation, routing based on global context, and cross-time normalization.

The synthetic modules are randomly initialized and audited without training. Input token IDs are sampled uniformly from the vocabulary; each pair shares the same sampled prefix and has independently sampled suffixes. These experiments test whether the implemented computation respects prefix invariance, rather than evaluating learned language-model quality. For each audited module, we compare outputs on the shared prefix. We then compute the LeakScore on prefix outputs using the default tolerance $\epsilon=10^{-5}$. A model is considered to fail the audit if the resulting LeakScore exceeds the tolerance, and to pass otherwise.

\subsection{Audited Modules}

Table~\ref{tab:synthetic_audit_main} summarizes the audited modules and their main results. The six non-causal variants all yield strictly positive LeakScore values, whereas all six causally repaired counterparts produce zero leakage under the test.

\begin{table}[t]
\centering
\small
\setlength{\tabcolsep}{4pt}
\renewcommand{\arraystretch}{1.15}
\caption{Six synthetic non-causal/causal implementation pairs. Score before and score after denote LeakScore for the non-causal implementation and its causal repair, respectively, under the same audit ($\epsilon=10^{-5}$). Scores are reported on each module's own output scale and should not be used to rank different modules.}
\label{tab:synthetic_audit_main}
\begin{tabularx}{\textwidth}{@{}l>{\raggedright\arraybackslash}X>{\raggedright\arraybackslash}Xrr@{}}
\toprule
Module & Non-causal operation & Causal repair & Before & After \\
\midrule
Pooling & Full-sequence mean & Token-wise routing & 0.0755 & 0.0000 \\
Pooling & Last-token broadcast & Prefix-mean routing & 0.3232 & 0.0000 \\
Conv1d & Symmetric padding & Left-only padding & 0.6625 & 0.0000 \\
Linear attention & Full-sequence KV aggregation & Prefix-only KV aggregation & 0.1168 & 0.0000 \\
Dynamic routing & Full-sequence context & Prefix-only context & 0.0033 & 0.0000 \\
Normalization & Full-sequence statistics & Prefix-only statistics & 0.0255 & 0.0000 \\
\bottomrule
\end{tabularx}
\end{table}

\subsection{Main Findings and Interpretation}

All six non-causal implementations fail the audit, while their paired causal repairs have zero discrepancy under the same conditions (Table~\ref{tab:synthetic_audit_main}). The failures span summary broadcast, temporal mixing, full-sequence KV aggregation, routing, and normalization, showing that the tested risk is not confined to pooling. LeakScore ranges from $0.0033$ to $0.6625$ across modules with different output scales, so these values should not be read as a cross-module severity ranking. The relevant comparison is within each implementation pair: every non-causal implementation lies well above the tolerance, and every causal repair is exactly zero.

These controlled comparisons connect the dependency criterion to concrete implementations without conflating causal validity with learned language-model quality. They also provide a check that the same behavioral test separates the evaluated failures from their causal counterparts. The following robustness checks vary the split, random seed, and tolerance; the real-model experiments then examine clean controls, injected violations, and runtime effects.

\subsection{Robustness Checks}

We further verified that these conclusions do not depend on a particular experimental setting. We consider three sources of variability: the prefix length used to define the split, the random seed controlling both model initialization and input sampling, and the tolerance used to decide pass or fail. Unless stated otherwise, the following checks use the same default configuration as the main experiments above (sequence length 32, prefix length 16, random seed 42, $\epsilon=10^{-5}$).

\paragraph{Varying the prefix length.}
Table~\ref{tab:robust_prefix_len} reports the LeakScore of each non-causal module under prefix lengths of $8$, $16$, $32$, and $64$. In order to accommodate prefix length 64, this check uses a longer sequence length of 80; this is why the numbers in Table~\ref{tab:robust_prefix_len} differ from those in Table~\ref{tab:synthetic_audit_main}, which used sequence length 32. Under all four split points, every non-causal module remains clearly above the tolerance, while every causally repaired variant remains at zero. The prefix-invariance failure is therefore not an artifact of a particular split position.

\begin{table}[t]
\centering
\caption{LeakScore of non-causal modules under different prefix lengths (sequence length 80, random seed 42, $\epsilon=10^{-5}$). All causally repaired variants remain at $0.0$ and are omitted for clarity.}
\label{tab:robust_prefix_len}
\begin{tabular}{lcccc}
\toprule
Module & 8 & 16 & 32 & 64 \\
\midrule
\texttt{noncausal\_pool\_mean} & 0.0484 & 0.0425 & 0.0328 & 0.0212 \\
\texttt{noncausal\_pool\_last} & 0.6819 & 0.5959 & 0.6969 & 0.6041 \\
\texttt{noncausal\_samepad\_conv1d} & 0.8203 & 0.7067 & 0.9781 & 0.9717 \\
\texttt{noncausal\_fullkv\_linear\_attn} & 0.1343 & 0.1134 & 0.0958 & 0.0711 \\
\texttt{noncausal\_global\_route} & 0.00421 & 0.00466 & 0.00359 & 0.00217 \\
\texttt{noncausal\_global\_norm} & 0.0358 & 0.0306 & 0.0199 & 0.0178 \\
\bottomrule
\end{tabular}
\end{table}

\paragraph{Varying the random seed.}
We repeated the audit with three random seeds ($0$, $1$, $2$) under the default configuration. Table~\ref{tab:robust_seed} summarizes the mean and population standard deviation of the LeakScore over the three seeds, together with the number of passing runs. All non-causal modules fail in all three runs ($0/3$), whereas every causally repaired variant passes in all three runs ($3/3$). The exact LeakScore values fluctuate with the seed, but the categorical conclusion is fully stable, and no module lands near the decision boundary.

\begin{table}[t]
\centering
\setlength{\tabcolsep}{4pt}
\caption{Seed stability of the audit over three random seeds (sequence length 32, prefix length 16, $\epsilon=10^{-5}$). All causally repaired variants achieve $3/3$ pass with LeakScore $0.0\pm0.0$.}
\label{tab:robust_seed}
\begin{tabular}{lcl}
\toprule
Module & Passed runs & LeakScore (mean $\pm$ std) \\
\midrule
\texttt{noncausal\_pool\_mean} & 0/3 & 0.0760 $\pm$ 0.0211 \\
\texttt{noncausal\_pool\_last} & 0/3 & 0.5104 $\pm$ 0.1420 \\
\texttt{noncausal\_samepad\_conv1d} & 0/3 & 0.6173 $\pm$ 0.0175 \\
\texttt{noncausal\_fullkv\_linear\_attn} & 0/3 & 0.1269 $\pm$ 0.0213 \\
\texttt{noncausal\_global\_route} & 0/3 & 0.00347 $\pm$ 0.00008 \\
\texttt{noncausal\_global\_norm} & 0/3 & 0.0316 $\pm$ 0.0104 \\
\bottomrule
\end{tabular}
\end{table}

\paragraph{Varying the tolerance.}
Finally, we repeated the audit with tolerance values of $10^{-6}$, $10^{-5}$, and $10^{-4}$. The pass/fail classification is identical across all three thresholds: every causally repaired variant passes and every non-causal module fails. Moreover, the weakest non-causal signal in our experiments remains above $10^{-4}$ (e.g., the non-causal global-routing variant at $\approx 0.0033$), which remains well above all three tested tolerances and the observed zero baseline of the repaired controls. The observed separation is therefore not an artifact of the tolerance choice.

\section{Real-Model Auditing}
\label{sec:real_model_audits}

The synthetic experiments establish that the audit has high specificity on controlled modules. We now ask whether the same test remains useful on real pretrained models, where implementations are black boxes and numerical noise is a practical concern. This section reports four additional layers of evidence: negative controls on trusted models, fault injection to establish sensitivity, cache and precision consistency across runtime paths, and layer-wise leakage localization of the first leak layer. Together they address a limitation of the toy setup: a behavioral test that works on synthetic modules might, in principle, either fire spuriously on real implementations or become difficult to interpret once runtime-path and precision effects introduce non-zero clean discrepancies.

\subsection{Experimental Setup}

We use three publicly released decoder-only models: the Transformers GPT-2 and Qwen2.5-0.5B, and the selective state-space model Mamba-130M~\cite{gu2024mamba}. The suffix-perturbation and localization audits use float32 inference with dropout disabled; the cache-consistency experiment separately evaluates float32, float16, and bfloat16.

The perturbation protocol follows the same fixed-prefix principle as the synthetic audit, with a separate evaluation grid. We set sequence length $T=64$, prefix lengths $t\in\{16,32,48\}$, random seeds $\{0,1,2\}$, and generate 8 suffix-perturbed pairs per combination of prefix length and seed, giving 72 pairs per model. Perturbations operate directly on token IDs, so the two inputs share exactly the same prefix tokens and differ only in the suffix, avoiding tokenizer boundary artifacts. For each pair we report the max and mean prefix-output discrepancy, the top-1 flip rate, the affected-token rate, and the pass/fail verdict under tolerance $\epsilon=10^{-5}$.

\subsection{Clean Negative Controls}

GPT-2, Qwen2.5-0.5B, and Mamba-130M each pass all 72 tested input pairs in float32 with $\epsilon=10^{-5}$; maximum and mean prefix-logit discrepancy and top-1 flip rate are all exactly zero. No false positives are observed under this same-path protocol. Exact zero does not imply that floating-point inference is generally noise-free: comparing different execution paths can yield non-zero discrepancies, as Section~\ref{sec:cache_precision} shows. The clean controls support the chosen tolerance for this protocol, rather than a universal threshold for other precisions or runtime paths.

\subsection{Fault Injection and Sensitivity}

A clean negative control only establishes specificity. To verify that the audit can actually detect a hidden leak when one is present, we inject a causal violation into the same models without touching the pretrained weights. Following the failure pattern found in the CDR case (Section~\ref{sec:cdr_case}), the wrapper first pools the full hidden-state sequence $h_{1:T}$ into a single summary $s=\mathrm{Mean}(h_{1:T})$, then produces a correction $\Delta z = \alpha\, g(s)$ broadcast back to every position, i.e.\ $z_i \leftarrow z_i + \Delta z$ for all $i$. This is exactly the sequence-level summary-to-broadcast pattern that our synthetic pooling routers instantiate, and it is applied identically to all three model families. We sweep the injection strength $\alpha\in\{0.01,0.05,0.10\}$; GPT-2 and Qwen are tested at all three strengths, and Mamba at $\alpha=0.05$, using the same 72-pair protocol.

Table~\ref{tab:real_injected} shows the results. Two findings stand out. First, sensitivity is high: at every tested non-zero injection strength, every evaluated model fails the audit and the pass rate drops from 1.0 to 0.0, while the affected-token rate stays at or above 0.993, i.e.\ the shared correction reaches essentially every prefix position. Second, the continuous diagnostics behave differently from the binary verdict. The max discrepancy increases monotonically with $\alpha$ (approximately linearly for Qwen: $1.6\times10^{-3}$, $7.9\times10^{-3}$, $1.6\times10^{-2}$), yet even the smallest tested injection $\alpha=0.01$ already fails the binary test. The audit is therefore both sensitive at the decision level and quantitative in the diagnostic scores, which is exactly the combination needed for the multi-metric diagnosis emphasized throughout this paper.

\begin{table}[t]
\centering
\small
\setlength{\tabcolsep}{4pt}
\caption{Fault-injection sensitivity with a sequence-level summary-broadcast wrapper and unchanged pretrained weights. Max diff and Norm. max are maxima over 72 pairs; JS divergence and affected-token rate are per-pair means, and pass rate is the fraction of passing pairs ($\epsilon=10^{-5}$).}
\label{tab:real_injected}
\resizebox{\textwidth}{!}{
\begin{tabular}{lcccccc}
\toprule
Model & $\alpha$ & Max diff & Norm. max & JS div & Affected-token rate & Pass rate \\
\midrule
GPT-2 & 0.01 & 2.1e-4 & 1.3e-6 & 1.9e-8 & 0.993 & 0.000 \\
GPT-2 & 0.05 & 1.0e-3 & 6.3e-6 & 1.8e-8 & 1.000 & 0.000 \\
GPT-2 & 0.10 & 2.1e-3 & 1.3e-5 & 2.0e-8 & 1.000 & 0.000 \\
Qwen2.5-0.5B & 0.01 & 1.6e-3 & 7.1e-5 & 1.1e-8 & 1.000 & 0.000 \\
Qwen2.5-0.5B & 0.05 & 7.9e-3 & 3.5e-4 & 9.9e-8 & 1.000 & 0.000 \\
Qwen2.5-0.5B & 0.10 & 1.6e-2 & 7.0e-4 & 3.9e-7 & 1.000 & 0.000 \\
Mamba-130M & 0.05 & 3.0e-3 & 3.7e-5 & 2.0e-8 & 1.000 & 0.000 \\
\bottomrule
\end{tabular}
}
\end{table}

\subsection{Cache and Precision Consistency}
\label{sec:cache_precision}

Cache consistency tests whether full-prefix evaluation and token-by-token decoding produce comparable logits for the same input and checkpoint. A discrepancy between these paths can arise from numerical precision, kernel differences, or cache implementation errors; it does not by itself demonstrate future-token leakage. We compare the two paths for GPT-2 and Qwen2.5-0.5B over a prefix of length 32 in float32, float16, and bfloat16. Mamba is outside the scope of this KV-cache experiment because it uses recurrent state rather than a Transformer key-value cache.

\begin{table}[H]
\centering
\small
\setlength{\tabcolsep}{4pt}
\caption{Cache-consistency audit between full-prefix forward and incremental decoding, summarized over the 32 prefix steps by the discrepancy distribution of the per-step maximum logit difference. The key calibration quantities are the clean-control P50/P95/max values for each runtime setting.}
\label{tab:cache_consistency}
\resizebox{\textwidth}{!}{
\begin{tabular}{llcccc}
\toprule
Model & Precision & P50 max diff & P95 max diff & Max diff & Top-1 agreement \\
\midrule
GPT-2 & fp32  & $1.9\times10^{-4}$ & $4.5\times10^{-4}$ & $7.6\times10^{-4}$ & $32/32$ \\
GPT-2 & fp16  & 0.125 & 0.25 & 0.625 & $32/32$ \\
GPT-2 & bf16  & 0.75 & 2.0 & 4.0 & $31/32$ \\
Qwen2.5-0.5B & fp32 & $2.5\times10^{-5}$ & $3.9\times10^{-5}$ & $8.1\times10^{-5}$ & $32/32$ \\
Qwen2.5-0.5B & fp16 & 0.0313 & 0.0476 & 0.1475 & $32/32$ \\
Qwen2.5-0.5B & bf16 & 0.2461 & 0.3703 & 0.5 & $32/32$ \\
\bottomrule
\end{tabular}
}
\end{table}

As Table~\ref{tab:cache_consistency} shows, the clean-control discrepancy distribution broadens sharply as precision is reduced. In fp32, the two runtime paths remain close but not numerically identical: the per-step P95 maximum discrepancy is $4.5\times10^{-4}$ for GPT-2 and $3.9\times10^{-5}$ for Qwen2.5-0.5B. In reduced precision the envelope becomes much wider, reaching a maximum of $4.0$ for GPT-2 in bf16 and $0.5$ for Qwen2.5-0.5B in bf16. Top-1 predictions remain fully consistent in every setting except GPT-2 bf16, where one of the 32 prefix steps has a different top-1 prediction. The implication is not that these clean checkpoints are architecturally non-causal, but that cache consistency is itself precision-sensitive: any behavioral threshold used across runtime paths must therefore be calibrated empirically on trusted clean controls, and should be reported together with the relevant precision/runtime setting rather than treated as universal. This operational dimension is complementary to, and in the spirit of, the cache-consistency checks used in recent work on causal correctness of sequence models \cite{fbs2026,torchlean2026}.

\subsection{Layer-wise Leakage Localization}

A failing verdict indicates detectable suffix-to-prefix influence, but does not identify where it first enters the network. To turn the audit from a binary decision into an actionable location, we compute a per-layer leakage profile. Let $h^{(\ell)}(x)$ denote the hidden state at the output of layer $\ell$. For a sequence pair $(x,x')$ sharing prefix $t$, the layer-$\ell$ leakage is
\[
L^{(\ell)}_{\max} = \max_{i \le t}\left\|h^{(\ell)}_i(x) - h^{(\ell)}_i(x')\right\|_\infty,
\]
and the \emph{first leak layer} is defined as
\[
\mathrm{FirstLeakLayer} = \min\left\{\ell : L^{(\ell)}_{\max} > \epsilon_\ell\right\},
\]
where layers are indexed from zero and $\epsilon_\ell=1.05\,c_\ell+10^{-8}$ is calibrated from clean controls under the same localization protocol, with $c_\ell$ the maximum clean discrepancy observed at layer $\ell$ across 24 calibration pairs. The threshold is set slightly above the empirically observed clean-control noise ceiling, so that floating-point variation is absorbed without masking genuine leakage. If no layer exceeds its threshold, we report that no first leak layer is detected for that input pair. To verify localization on the three public models, we inject a summary-broadcast perturbation into the hidden-state output of layer 3 with $\alpha=0.05$ and check whether this layer is recovered on one representative input pair ($T=64$, $t=32$) per model. These profiles are not maxima over the 72-pair output audit. The representative pair is also among the clean calibration pairs, so this check is not an independent hold-out evaluation. This hidden-state intervention differs from the output-logit wrapper used in the sensitivity experiment; their discrepancy magnitudes are not directly comparable.

As Table~\ref{tab:localization} shows, the injected layer is recovered in all three tested architectures: on clean models no layer exceeds its threshold, and after injection the audit identifies layer 3 as the first monitored layer with detectable suffix influence. The final-layer discrepancy is larger than at the injection layer. These results support using layer-wise profiles to guide implementation inspection under the evaluated conditions. The same diagnostic is applied to the CDR checkpoints in Section~\ref{sec:cdr_case}, where the first detectable leak occurs at the bottom of the network.

\begin{table}[H]
\centering
\caption{Hidden-state localization on one representative pair per model ($t=32$) after injection at zero-based layer 3 ($\alpha=0.05$). Final-layer discrepancy is measured at the last monitored layer.}
\label{tab:localization}
\begin{tabular}{lccc}
\toprule
Model & Layers & Inject $\rightarrow$ FirstLeakLayer & Final-layer discrepancy \\
\midrule
GPT-2 & 12 & $3{\rightarrow}3$ & 3.47 \\
Qwen2.5-0.5B & 24 & $3{\rightarrow}3$ & 0.86 \\
Mamba-130M & 24 & $3{\rightarrow}3$ & 18.15 \\
\bottomrule
\end{tabular}
\end{table}

\begin{figure}[H]
\centering
\includegraphics[width=\textwidth]{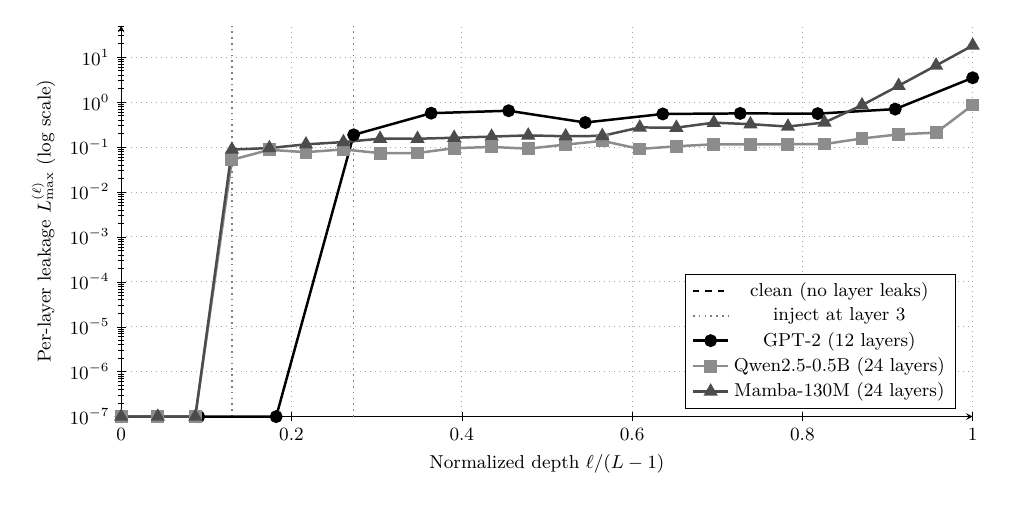}
\caption{Layer-wise hidden-state discrepancy after fault injection at zero-based layer 3 ($\alpha=0.05$). Dotted lines indicate the injection depth. Zero discrepancies are displayed at $10^{-7}$ on the logarithmic axis.}
\label{fig:e4_propagation}
\end{figure}

Figure~\ref{fig:e4_propagation} shows the full per-layer profile. For all three models the leakage is exactly zero before the injection point, jumps to a non-zero value at layer 3, and then propagates through subsequent layers with local decreases as well as increases; the final-layer discrepancy exceeds the discrepancy at the injection layer in every model. The clean models remain at zero in every layer. Because the models differ in depth (12 vs.\ 24 layers), the same absolute layer 3 appears at different normalized depths in the figure ($3/11$ for GPT-2 vs.\ $3/23$ for the others); the recovered FirstLeakLayer is nonetheless the exact absolute layer index in every case.

\section{Case Study: False Architectural Gains in CDR-AttnRes}
\label{sec:cdr_case}

We examine the main empirical case from our architectural exploration: CDR-AttnRes, a residual-routing decoder variant developed in follow-up experiments inspired by Attention Residuals \cite{kimiteam2026attentionresiduals}. In this case, an anomalously low teacher-forced validation loss does not indicate an architectural improvement: the variants also exhibit detectable prefix leakage and markedly degraded free-running generation; layer-wise localization and implementation inspection trace the leakage to a non-causal query pathway; and causal repairs eliminate both the leakage and the large apparent advantage.

We use a compact naming scheme for the checkpoints audited in this section. \emph{Baseline} denotes the standard decoder, and \emph{Static AttnRes} uses learned base routing queries without the content-conditioned query correction introduced by CDR; its depth-wise mixing weights remain input-dependent (AttnRes abbreviates Attention Residuals \cite{kimiteam2026attentionresiduals}). The 12-layer pre-fix variants share the CDR-AttnRes routing and rank $r=64$, and differ in the sequence-level pooling used to produce the query correction: \emph{CDR-mean}, \emph{CDR-gated}, and \emph{CDR-last} correspond to mean, gated, and last-token pooling, respectively. Causally repaired reimplementations carry a \emph{-fix} suffix (\emph{CDR-token-fix} and \emph{CDR-prefix-mean-fix}). A name without a \emph{24L} suffix denotes the 12-layer configuration.

\subsection{Anomalous Training and Validation Behavior}

The first sign of the problem was an unusually large reduction in validation loss. In a 24-layer configuration, the baseline model achieved a validation loss of $3.5958$, whereas the pre-fix CDR-last 24L variant reached $1.4259$. This large gap motivated a causal audit before interpreting the lower loss as an architectural improvement.

The same pattern also appeared in 12-layer experiments. We therefore use the anomalous 24-layer checkpoint as the trigger case, while the 12-layer variants provide the controlled and repair comparisons used to test whether the same mechanism recurs and can be causally repaired. Several pre-fix variants based on sequence-level pooling all achieved much lower validation losses than the trusted controls, \emph{Baseline} and \emph{Static AttnRes}. However, these gains were not accompanied by correspondingly credible autoregressive behavior.

\subsection{Direct Audit with Prefix Invariance and LeakScore}

We next audited the real checkpoints using the same prefix-invariance test introduced earlier. For each checkpoint, we used sequence length $T=64$, prefix lengths $t\in\{16,32,48\}$, seeds $\{0,1,2\}$, and eight suffix perturbations per combination, giving 72 input pairs with identical prefixes and different suffixes. We compared the prefix logits with tolerance $\epsilon=10^{-5}$. At the checkpoint level, we report the maximum LeakScore over the 72 pairs; mean absolute discrepancy, top-1 flip rate, and affected-token rate are arithmetic means of their per-pair values. Thus, rates give equal weight to each pair rather than pooling prefix tokens across different prefix lengths. For the anomalous pre-fix 24-layer checkpoint, the audit fails decisively: the maximum prefix-output difference reaches $34.22$, and the mean per-pair top-1 flip rate is $95.1\%$.

The checkpoint-level failure establishes detectable suffix-to-prefix influence before localization or implementation inspection. It links the validation-loss anomaly to a direct violation of the causal contract; Table~\ref{tab:leakgain} reports the corresponding audit results across the core checkpoints.

\subsection{Layer-wise Localization of the CDR Leak}

We also applied the same layer-wise localization criterion used in Section~\ref{sec:real_model_audits} to the CDR checkpoints, using causal-safe references for the per-layer thresholds: the token-wise repaired model for 12-layer pre-fix variants and the 24-layer baseline for the 24-layer pre-fix variant. Controls and repaired variants use self-calibration. The localization profiles use a representative input pair with sequence length 64 and prefix length 32, while the per-layer thresholds are calibrated on 24 pairs. The representative pair is included among the calibration inputs, so this is not an independent hold-out evaluation. These profiles are not maxima over the 72-pair checkpoint audit. Table~\ref{tab:cdr_localization} reports the representative trusted and pre-fix variants. Every trusted control has no first leak layer and zero final-layer discrepancy. By contrast, every pre-fix variant shown in the table, including the 24-layer checkpoint, begins to leak immediately at layer 0. The discrepancy then propagates through subsequent layers with local decreases as well as increases; in every leaking variant, the final-layer discrepancy exceeds that at layer 0.

\begin{table}[H]
\centering
\caption{CDR hidden-state localization on a representative input pair with prefix length 32. Final-layer discrepancy is measured at the last monitored layer; -- denotes no detected leaking layer. Unmarked variants are 12-layer.}
\label{tab:cdr_localization}
\small
\begin{tabular}{lcc}
\toprule
Variant & FirstLeakLayer & Final-layer discrepancy \\
\midrule
Baseline & -- & 0.00 \\
Static AttnRes & -- & 0.00 \\
\midrule
CDR-gated & $0$ & 2.20 \\
CDR-mean & $0$ & 3.55 \\
CDR-last & $0$ & 4.64 \\
CDR-last 24L & $0$ & 16.96 \\
\bottomrule
\end{tabular}
\end{table}

\subsection{Generation Diagnostics of Controls and Leaking Checkpoints}

To assess whether the audit failure accompanies degraded generation, we evaluated the checkpoints under a fixed open-ended generation protocol. For each variant, we generated 50 continuations from fixed prompts taken from the OpenWebText validation set (50 prompts derived from the first 32 GPT-2 tokens of each selected document, with whitespace normalized before generation, $300$ new tokens, temperature $0.8$, top-$k$ $200$, random seed $1337$), and measured reference-model perplexity $\mathrm{ppl}_{\mathrm{ref}}$ (sliding-window perplexity of the GPT-2 medium reference LM; definition in Section~\ref{sec:method}), sampling-distribution entropy $H$, and lexical diversity and repetition statistics. Table~\ref{tab:generation} reports prompt-averaged values. As defined in Section~\ref{sec:method}, reference perplexity and lexical statistics use the saved prompt-inclusive text, whereas entropy is averaged over generated steps. The validation-loss column records checkpoint validation results and is not an average over the generation prompts. Per-checkpoint 95\% confidence intervals for mean reference perplexity are computed by resampling the 50 prompts 2000 times; the interval envelopes across checkpoint groups are summarized below.

\begin{table}[t]
\centering
\caption{Validation loss and generation diagnostics for controls and pre-fix CDR checkpoints. Generation metrics are averaged over 50 prompts, with one continuation per prompt; validation loss is the checkpoint validation result. Unmarked variants are 12-layer.}
\label{tab:generation}
\begin{tabular}{lccccc}
\toprule
Variant & Val loss & $\mathrm{ppl}_{\mathrm{ref}}$ & Entropy & Distinct-2 & Repeat-3 \\
\midrule
Static AttnRes & 3.6676 & 17.2 & 1.64 & 0.804 & 0.174 \\
Baseline & 3.6900 & 19.8 & 1.85 & 0.808 & 0.153 \\
Baseline 24L & 3.5958 & 16.8 & 1.90 & 0.827 & 0.139 \\
\midrule
CDR-gated & 2.6015 & 198.5 & 2.31 & 0.822 & 0.136 \\
CDR-mean & 2.8473 & 299.9 & 2.80 & 0.725 & 0.198 \\
CDR-last & 2.5488 & 695.5 & 2.87 & 0.784 & 0.134 \\
CDR-last 24L & 1.4259 & 700.5 & 2.13 & 0.849 & 0.086 \\
\bottomrule
\end{tabular}
\end{table}

Table~\ref{tab:generation} shows a clean separation. Every trusted control has $\mathrm{ppl}_{\mathrm{ref}}$ between $16.8$ and $19.8$, whereas every pre-fix variant lies between $198.5$ and $700.5$---roughly a 10--42$\times$ gap with no overlap. The per-prompt uncertainty estimates reinforce the separation: the 95\% bootstrap confidence intervals of the trusted controls (whose individual interval endpoints span [15.2, 22.8]) and the leaking variants (whose endpoints span [169.0, 839.7]) do not overlap. Sampling entropy provides a second descriptive separation: controls occupy $1.64$--$1.90$, and leaking variants $2.13$--$2.87$. The same checkpoints that fail the causal audit have much higher reference perplexity and sampling entropy during generation.

Lexical diversity alone does not capture the observed degradation. The pre-fix 24-layer variant has the highest Distinct-2 ($0.849$) and lowest Repeat-3 ($0.086$) in Table~\ref{tab:generation}, yet also the highest reference perplexity ($700.5$). These lexical scores therefore do not establish generation quality. Reference perplexity and the illustrative samples in ~\ref{app:e6_samples} support the diagnosis of off-target generation, while sampling entropy describes the distribution from which tokens are selected. The small perplexity difference between the two last-pooling checkpoints should not be interpreted as a reliable ordering, given their overlapping confidence intervals.

\subsection{Root Cause: Sequence-Level Pooling in Dynamic Query Construction}

To explain why the checkpoint fails the audit and why localization points to the very bottom of the network, we inspected the implementation paths that inject auxiliary signals into the backbone, which led us to the dynamic query pathway. Here, the query controls depth-wise routing: at each token position, it scores residual sources from different depths, and a softmax over those sources determines their weighted mixture. This routing is applied before both the attention and MLP sublayers; the query is distinct from the token-to-token query in masked self-attention. In the pre-fix implementation, a sequence-level summary was first computed from the entire hidden-state sequence, using mean pooling, last-token pooling, or related summary operations. The hidden-state sequence of shape $B\times T\times D$ was pooled along the time axis to a vector of shape $B\times D$, and the resulting summary was then projected into a shared dynamic query correction $\Delta q_{\mathrm{shared}}$ that was broadcast back to all token positions. The same correction signal therefore influenced prefix positions as well as later ones.

This design is non-causal because the sequence-level summary already contains information from suffix tokens. As a result, even though the masked self-attention path remained causally correct, prefix positions were indirectly affected by $x_{t+1:T}$ through the auxiliary dynamic-query route. This mechanism also explains the localization result above: because the shared correction is injected immediately into the backbone computation, the first detectable leak appears at the bottom of the network and is then amplified by subsequent layers.

\subsection{Repair Validation}

We finally asked whether a causality-preserving reimplementation would remove both the audit failure and the apparent empirical gain. The repair replaces the shared full-sequence summary with a position-dependent causal summary of the current hidden states $h_{1:T}$. Token-wise repair uses $s_i=h_i$, whereas prefix-mean repair uses $s_i=i^{-1}\sum_{j=1}^{i}h_j$. Both retain the low-rank correction $\Delta q_i=W_{\mathrm{up}}\,\mathrm{GELU}(W_{\mathrm{down}}s_i)$ and add it to the learned base routing query at position $i$, so the correction has no access to future positions. The two repaired variants were trained from scratch using the shared training recipe in ~\ref{app:training}; they are separate checkpoints rather than inference-time switches applied to a pre-fix checkpoint. Table~\ref{tab:cdr_repair} summarizes their results. Under both token-wise and prefix-mean repair, the LeakScore returns to zero, no first leak layer is found, the validation loss is close to that of the trusted controls, and the generation perplexity ($15.3$ and $18.1$) returns to the same scale as the trusted controls ($16.8$--$19.8$) while remaining an order of magnitude below every leaking variant ($198.5$--$700.5$).

\begin{table}[t]
\centering
\caption{Audit and evaluation results for the two independently trained 12-layer causal repairs. -- denotes no detected leaking layer. Both repairs pass the audit and have validation loss and reference perplexity on a similar scale to the controls.}
\label{tab:cdr_repair}
\begin{tabular}{lcccc}
\toprule
Variant & Val loss & LeakScore & FirstLeakLayer & $\mathrm{ppl}_{\mathrm{ref}}$ \\
\midrule
CDR-token-fix & 3.6766 & 0.0 & -- & 15.3 \\
CDR-prefix-mean-fix & 3.6731 & 0.0 & -- & 18.1 \\
\bottomrule
\end{tabular}
\end{table}

\begin{figure}[!htbp]
\centering
\includegraphics[width=\textwidth]{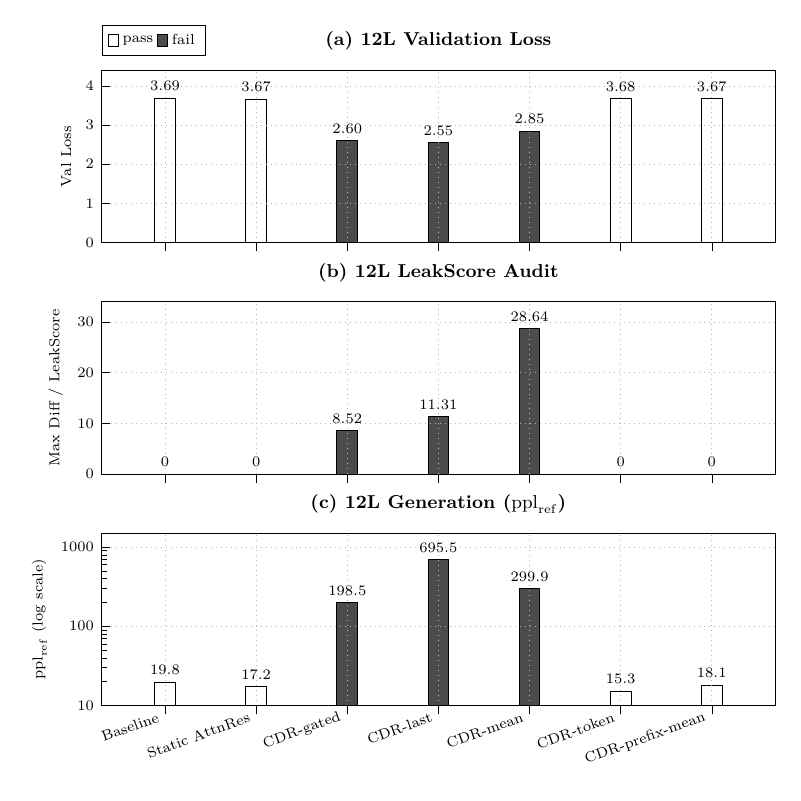}
\caption{Diagnostics for seven 12-layer checkpoints in a shared order: (a) validation loss, (b) checkpoint-level LeakScore, and (c) mean reference perplexity over 50 prompts. Open and shaded bars denote passing and failing checkpoints, respectively. Panel (c) uses a logarithmic axis with untransformed perplexity labels. CDR-token and CDR-prefix-mean abbreviate the two repaired variants.}
\label{fig:cdr_case_study}
\end{figure}

Figure~\ref{fig:cdr_case_study} compares all seven 12-layer checkpoints using stacked panels sharing a common horizontal axis: validation loss (a), checkpoint-level LeakScore (b), and generation perplexity (c). The leaking variants combine lower validation loss with higher generation perplexity, whereas the repaired variants have zero leakage and return to the control scale. Table~\ref{tab:cdr_localization} provides layer-wise localization, and Table~\ref{tab:cdr_repair} reports the repair results numerically.

\subsection{Quantitative Consistency Between Leakage and Apparent Gain}

Beyond the focused anomaly and repair comparisons above, we also asked whether the leak--gain relationship holds over the nine core checkpoints rather than only the representative examples above. We therefore ran the same audit on the full set of nine core CDR variants. Table~\ref{tab:leakgain} reports validation loss, maximum leakage, top-1 flip rate, and the audit verdict for each.

\begin{table}[t]
\centering
\caption{Prefix-invariance audit on nine core CDR checkpoints ($\epsilon=10^{-5}$). LeakScore is the maximum over 72 pairs; top-1 flip rate is the mean of per-pair rates. Controls and repairs pass on all tested pairs, while all pre-fix variants fail. Unmarked variants are 12-layer.}
\label{tab:leakgain}
\begin{tabular}{lcccc}
\toprule
Variant & Val loss & LeakScore & Top-1 flip rate & Verdict \\
\midrule
Baseline & 3.6900 & 0.0 & 0.0 & pass \\
Static AttnRes & 3.6676 & 0.0 & 0.0 & pass \\
Baseline 24L & 3.5958 & 0.0 & 0.0 & pass \\
\midrule
CDR-token-fix & 3.6766 & 0.0 & 0.0 & pass \\
CDR-prefix-mean-fix & 3.6731 & 0.0 & 0.0 & pass \\
\midrule
CDR-gated & 2.6015 & 8.52 & 0.483 & fail \\
CDR-last & 2.5488 & 11.31 & 0.509 & fail \\
CDR-mean & 2.8473 & 28.64 & 0.720 & fail \\
CDR-last 24L & 1.4259 & 34.22 & 0.951 & fail \\
\bottomrule
\end{tabular}
\end{table}

Across the nine checkpoints, lower validation loss is associated with greater leakage (Pearson $r=-0.868$ for LeakScore; full descriptive statistics are provided in the supplementary materials). This pattern is largely a separation between the five passing controls/repairs and four failing pre-fix checkpoints. Within the four leaking checkpoints alone, the association is weaker and not statistically significant (Pearson $r=-0.559$, $p=0.441$). These related checkpoints are not independent experimental replications, and the leaking variants differ in pooling mode and depth while sharing rank $r=64$. The small sample and these differences preclude a dose--response interpretation. Table~\ref{tab:leakgain} supports the case-level observation that the apparent low-loss advantage accompanies audit failure; it does not establish that leakage magnitude determines validation loss.

Together, the audit, generation diagnostics, implementation inspection, and 12-layer repair comparisons support the interpretation that the large pre-fix gain depends on future-token access. The independently trained repairs do not isolate every effect of the implementation change, and we do not report a repaired 24-layer checkpoint. Within these boundaries, the joint disappearance of audit failure and anomalously low loss means that the original result cannot be treated as evidence of a valid autoregressive improvement.

\subsection{Implications of the Case Study}

The practical implication of this case is to audit the complete computation before interpreting a lower teacher-forced validation loss as an architectural improvement. Layer-wise localization can guide implementation inspection, and independently trained causal repairs can test whether the apparent advantage persists under a valid dependency structure.

\section{Guidelines and Limitations}
\label{sec:discussion}

\subsection{Causality-Safe Design Guidelines}

The implementation question is what information can influence an output at position $i$. Every contributing summary, state update, routing decision, and normalization statistic must respect the prefix boundary. Table~\ref{tab:synthetic_audit_main} gives concrete repairs: token-local or prefix-only summaries, left-padded convolution, prefix-KV aggregation, and prefix-consistent routing and normalization. A causal attention mask cannot compensate for future access elsewhere in the computation.

The CDR case gives a specific example: the query controls mixing across depth, yet the summary used to construct it mixes across time. Restricting the token-to-token attention pattern therefore leaves this auxiliary path unchecked. Auditing should cover such control signals as well as the primary sequence-mixing operator, with clean controls matched to the execution conditions. Prefix-invariance tests can reveal violations and guide inspection, while a passing result remains limited to the tested inputs and tolerance.

\subsection{Limitations}

Our results should be interpreted with several limitations in mind. First, the prefix-invariance audit used here should be viewed as a necessary practical check rather than a formal proof of causality. Passing the test provides evidence that a model does not exhibit detectable suffix-to-prefix influence under the evaluated perturbations and tolerance, but it does not establish that every possible execution path or implementation detail is causally correct in a formal verification sense. A failure exceeding a calibrated tolerance under controlled evaluation is evidence of suffix-to-prefix influence; attributing it to a hidden auxiliary path requires implementation or localization evidence. New architectural components should therefore undergo implementation review as well as behavioral testing.

Second, although we study several representative families of auxiliary modules, our coverage is not exhaustive. In particular, we do not claim to cover all state-space models, all mixture-of-experts designs, or the full range of hybrid architectures combining routing, compression, recurrence, retrieval, and state updates. Some SSM or MoE variants may be causally safe by construction, whereas others may contain subtler non-causal paths than the ones examined here. Extending the audit to a broader set of modern sequence architectures remains an important direction for future work.

Third, the synthetic experiments are intentionally simplified in order to isolate causal structure from unrelated optimization effects. For that reason, they should be interpreted as controlled behavioral audits rather than as task-level benchmark evaluations of architectural quality. The purpose of these experiments is to test whether a module can leak future information, not to provide a comprehensive comparison of model performance across datasets and training regimes.

Finally, the CDR-AttnRes case study offers a realistic and informative example, but it is still a single concrete case from our own architectural exploration. Additional case studies on independent codebases and publicly released implementations would further strengthen the empirical scope of the paper. We therefore regard the present results as evidence for the importance of causal auditing, rather than as a claim that every adaptive sequence model in the literature is affected by hidden future-token leakage.

\section{Conclusion}
\label{sec:conclusion}

Hidden future-token leakage can create false architectural gains even when the causal attention mask is correct. In the CDR-AttnRes case, anomalously low teacher-forced validation loss co-occurs with detectable prefix leakage and substantially worse reference-model perplexity during free-running generation. Layer-wise diagnostics and implementation inspection identify a full-sequence summary broadcast through the depth-routing query path. Two independently trained 12-layer causal repairs remove detectable leakage, return validation loss and reference perplexity to the control scale, and remove the large apparent advantage.

The main empirical finding is that, with a correct causal attention mask, hidden future-token leakage is still sufficient to produce a large performance advantage that invites interpretation as an architectural improvement. That advantage disappears under causal repair and is therefore not a valid autoregressive improvement. The established prefix-invariance contract, LeakScore, complementary metrics, and FirstLeakLayer provide diagnostic evidence for the case. Six synthetic implementation pairs, clean pretrained controls, controlled fault injection, and cache/precision checks support interpretation of those diagnostics under the tested conditions. They do not constitute a proof of causality for all inputs or implementations.

The practical implication is to check the complete dependency structure before treating lower teacher-forced loss as architectural progress. Pooling, temporal mixing, state aggregation, routing, and cross-time normalization can introduce future-token access outside masked attention. Combining causal checks with free-running evaluation, root-cause inspection, and repair validation can help distinguish a valid autoregressive improvement from an apparent gain enabled by an invalid dependency path.

\appendix
\setcounter{table}{0}
\section{Reproducibility}
\label{sec:appendix}

\subsection{CDR Training Configuration}
\label{app:training}

All CDR checkpoints were trained with the shared configuration in Table~\ref{tab:training}; both the 12-layer and the 24-layer variants use the same hyper-parameters.

\begin{table}[H]
\centering
\caption{Shared CDR training configuration. Parameter counts are unique trainable parameters; the input embedding and the output head share a single weight matrix, counted once. CDR adds rank-$r$ down- and up-projections for dynamic routing-query corrections in the residual aggregation paths before both the attention and MLP sublayers.}
\label{tab:training}
\small
\begin{tabular}{lp{8.4cm}}
\toprule
Data & OpenWebText, GPT-2 tokenizer (vocabulary padded from 50257 to 50304), block size 512 \\
Optimizer & AdamW ($\beta=(0.9,0.95)$, weight decay 0.1, gradient clip 1.0), dropout 0.0, no bias \\
Learning rate & $3\times10^{-4}$, cosine decay to $3\times10^{-5}$, warmup 1000 iterations \\
Iterations & 20{,}000 \\
Effective batch & 64 sequences/step (batch size 4 per device, gradient accumulation 16/2 GPUs $= 8$ per device, 2 devices) \\
Precision & bfloat16 \\
Training seed & 1337 (all nine core checkpoints) \\
Checkpoint selection & best validation loss (evaluation every 1000 iterations, 100 evaluation iterations) \\
Model sizes & 12 layers, 8 heads, 512-dimensional embedding: 63.8M parameters (baseline), 65.4M with CDR ($r=64$); 24 layers, 8 heads, 512-dimensional embedding: 101.5M (baseline), 104.7M with CDR ($r=64$) \\
\bottomrule
\end{tabular}
\end{table}

\subsection{Auditing Protocols}
\label{app:protocols}

Public-model audit conditions are specified in Section~\ref{sec:real_model_audits}; CDR audit and generation conditions are specified in Section~\ref{sec:cdr_case}, with metric definitions in Section~\ref{sec:method}. The supplementary materials provide per-experiment result files, the fixed prompt set, saved continuations, and the checkpoint-to-implementation mapping. The nine core checkpoints use training seed 1337; auxiliary rank and seed explorations are not part of the core checkpoint comparison.

\subsection{Representative Generation Samples}
\label{app:e6_samples}

We show the first 80 whitespace-delimited units of the saved continuation for the same prompt ($\mathrm{idx}=0$) from three 12-layer checkpoints: Baseline, pre-fix CDR-last, and repaired CDR-token-fix. Wording and generated errors are preserved; line breaks and quotation marks are normalized for typesetting, and [\ldots] marks truncation. The displayed \texttt{\textless{}|endoftext|\textgreater{}} marker is present in the saved baseline output. These illustrative excerpts do not establish factual accuracy or replace the aggregate diagnostics in Table~\ref{tab:generation}. Full outputs for all 50 prompts, including the 24-layer variants, are provided in the supplementary materials.

\paragraph{Shared prompt}
\begin{quote}\small
Reporting from Perry, Iowa --- Defending himself against charges of isolationism, Republican presidential candidate Ron Paul told voters in Iowa on Thursday that western sanctions against Iran are
\end{quote}

\paragraph{Baseline (12L control)}
\begin{quote}\small
too high. ``Don't believe me when you get to war,'' he said by phone. ``This is something that the government is supposed to protect,'' he said.\textless{}|endoftext|\textgreater{}There's a reason for the Trump administration's decision to use climate "perform" heathens to decry the risks of climate change under the President-elect Donald Trump. "Because the climate change ban on its own will not be renewed," said Kelly "I am sure that it will be renewed as soon as possible.'' How much does warming [\ldots]
\end{quote}

\paragraph{CDR-last (12L pre-fix)}
\begin{quote}\small
self un un state un low un low low heavy self heavy un highly un heavy un self un self un heavy inf self un un highly low heavy deep heavy kind low large form form un particularly large very largely very small low un high large inf human super small heavy low heavily small large small un heavy large single super completely large single un large form small traditional large large large single super large large large super low [\ldots]
\end{quote}

\paragraph{CDR-token-fix (12L repair)}
\begin{quote}\small
too sweeping but he continued with an comments he made earlier this month when he asked reporters to listen to him. ``His statements were going to be something, and it was a hard thing to get him to do it, to attack people,'' Paul said. ``He was, after all, a very, very good friend. He had a really good message. But he knew he thought it was important to have a back-up.'' Paul, a senator from Virginia, said he has [\ldots]
\end{quote}

\section*{Funding}
This paper was supported in part by the China Guangxi Science and Technology Plan Project under grant AD23026096, Chile CONICYT FONDECYT Regular under grant 1181809, and Chile CONICYT FONDEF under grant ID16I10466.

\section*{Data and Code Availability}

The accompanying supplementary materials contain the auditing and generation code, saved results for the synthetic and public-model experiments, CDR audit and generation summaries, the fixed OpenWebText prompt set, raw continuations, descriptive correlation statistics, and checkpoint-to-implementation mappings. Model checkpoints are not included because of storage constraints. Checkpoint inference requires the corresponding weights; retraining additionally requires preprocessed OpenWebText data and the training environment. Public-model audits and reference scoring require the relevant public and reference models. Paths to external checkpoints and datasets must be configured before running the corresponding experiments; exact numerical agreement across environments is not guaranteed.

\section*{Declaration of generative AI and AI-assisted technologies in the manuscript preparation process}

During the preparation of this work, the authors used DeepSeek-V4-Flash in order to improve the clarity and readability of the manuscript text. After using this tool, the authors reviewed and edited the content as needed and take full responsibility for the content of the published article.

\bibliographystyle{elsarticle-num}
\setlength{\bibsep}{3pt}
\bibliography{refs}

\end{document}